\documentclass[twoside]{article}
\usepackage[accepted]{aistats2016_s}

\usepackage{amsmath,verbatim,natbib,epsfig,enumitem,graphicx}
\usepackage{amssymb}

\newcommand{\cind}{\perp\hspace*{-1.35ex}\perp}
\usepackage{graphicx}
\usepackage[linesnumbered]{algorithm2e}
\usepackage{url}

\bibpunct{(}{)}{;}{a}{,}{,}

\begin{document}

\runningtitle{Learning Instrumental Variables with Non-Gaussianity Assumptions}

%

\twocolumn[

\aistatstitle{Learning Instrumental Variables with Non-Gaussianity Assumptions:\\ Theoretical Limitations and Practical Algorithms}

\aistatsauthor{Ricardo Silva \And Shohei Shimizu}

\aistatsaddress{
Department of Statistical Science / CSML\\University College London\\{\tt ricardo@stats.ucl.ac.uk} \And 
Institute of Scientific and Industrial Research\\Osaka University\\{\tt sshimizu@ar.sanken.osaka-u.ac.jp}} 

]

\begin{abstract}
  Learning a causal effect from observational data is not
  straightforward, as this is not possible without further
  assumptions. If hidden common causes between treatment $X$ and
  outcome $Y$ cannot be blocked by other measurements, one possibility
  is to use an instrumental variable. In principle, it is possible
  under some assumptions to discover whether a variable is
  structurally instrumental to a target causal effect $X \rightarrow
  Y$, but current frameworks are somewhat lacking on how general these
  assumptions can be. A instrumental variable discovery problem is
  challenging, as no variable can be tested as an instrument in
  isolation but only in groups, but different variables might require
  different conditions to be considered an instrument. Moreover,
  identification constraints might be hard to detect statistically. In
  this paper, we give a theoretical characterization of instrumental
  variable discovery, highlighting identifiability problems and
  solutions, the need for non-Gaussianity assumptions, and how they
  fit within existing methods.
\end{abstract}

\section{CONTRIBUTION}

\begin{figure}[top]
\label{fig:iv}
\begin{center}
\includegraphics[height=2.1cm]{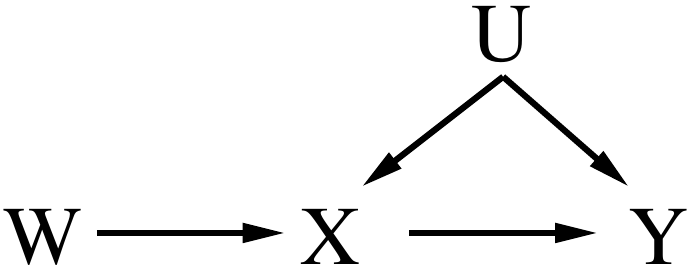}
\end{center}
\caption{A graph illustrating a possible IV structure. $X$ and $Y$ have an unmeasured confounder
$U$. $W$ is an instrument as it is unconfounded with $Y$, has no direct effect of it, and
causes $X$. In this paper, variables named ``$U$'' will denote unmeasured variables.}
\end{figure}

Consider a linear graphical causal model \citep{sgs:00, pearl:00}, where given a 
directed acyclic graph (DAG) $\mathcal G$, we define a joint distribution in terms
of conditional relationships between each variable $V_i$ and its given {\it parents}
in $\mathcal G$:
\begin{equation}
V_i = \sum_{V_j \in par_{\mathcal G}(i)}\lambda_{ij}V_j + e_i.
\label{eq:sem}
\end{equation}
\noindent That is, each random variable $V_i$ is also a vertex in $\mathcal G$,
where $par_{\mathcal G}(i)$ are the parents of $V_i$ in $\mathcal G$
and $e_i$ is an independent error term. Equation (\ref{eq:sem}) is
called a {\it structural equation} in the sense it encodes a
relationship that remains stable under a {\it perfect intervention} on
other variables. Using the notation of \cite{pearl:00}, we use the
index ``$do(V_k = v_k)$'' to denote the regime under which some
variable $V_k$ is fixed to some level $v_k$ by an external agent.  If
$V_k$ is a parent of $V_i$, the {\it differential causal effect} of
$V_k$ on $V_i$ is given by:
\begin{equation}
\frac{\partial E[V_i\ |\ do(V_k = v_k)]}{\partial v_k} = \lambda_{ik}.
\end{equation}
\noindent Each $\lambda_{ik}$ will be referred to as a {\it structural
coefficient}. Our goal is to estimate the differential causal effect of some
treatment $X$ on some outcome $Y$ from observational data. If the
common hidden causes of these two variables can be blocked by other
observable variables, a formula such as the back-door adjustment of
\cite{pearl:00} or the Prediction Algorithm of \cite{sgs:00} can
be used to infer it.  In general, unmeasured confounders of $X$ and
$Y$ might remain unblocked. In linear models, a possibility is to use
an {\it instrumental variable} (or {\it instrument}, or {\it IV}): some observable variable
$W$ that is not an effect of either $X$ and $Y$, it is unconfounded
with $Y$, and has no direct effect on $Y$. Figure \ref{fig:iv}
illustrates one possible DAG containing an instrument, with further details
in the next Section.

\begin{figure*}[top]
\begin{center}
\begin{tabular}{cccc}
\includegraphics[height=2.1cm]{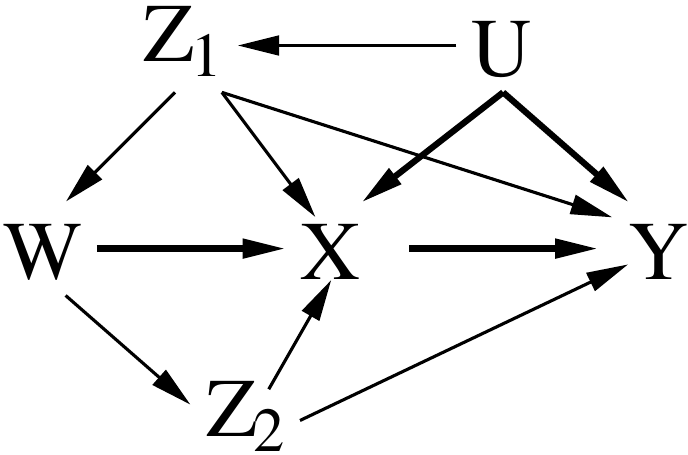} &
\includegraphics[height=2.1cm]{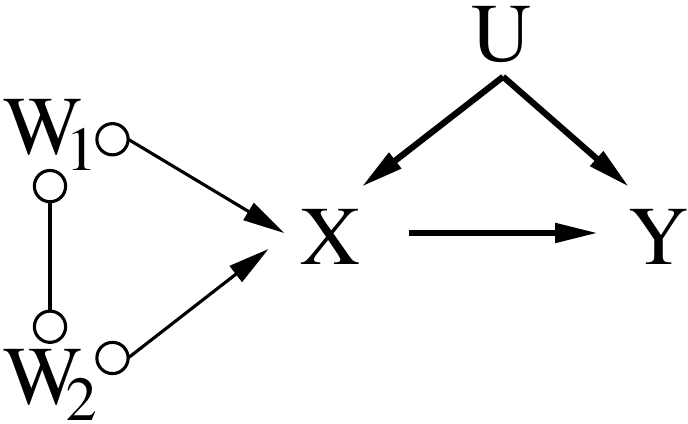} &
\includegraphics[height=2.1cm]{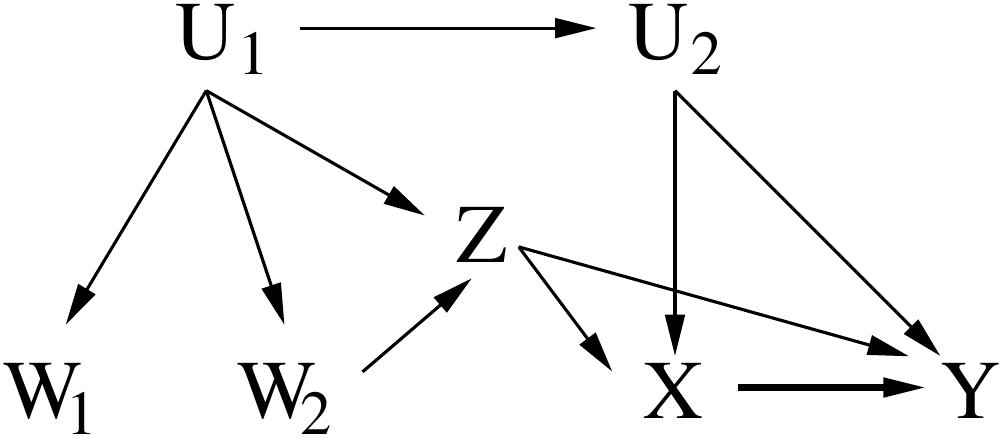} &
\includegraphics[height=2.1cm]{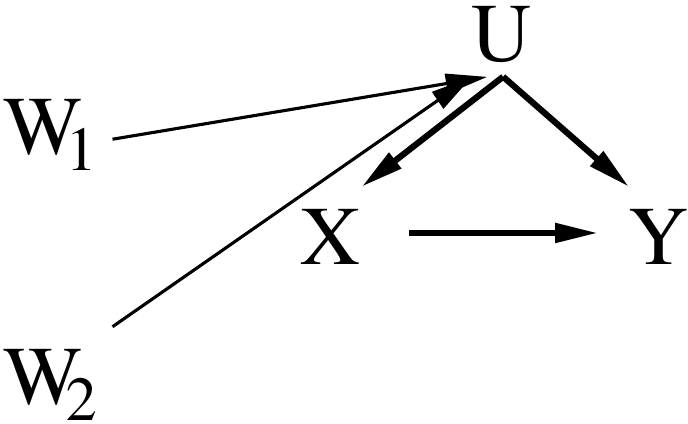}\\
(a) & (b) & (c) & (d)\\
\end{tabular}
\end{center}
\caption{(a) Variable $W$ is an instrument for the relation $X \rightarrow Y$ conditioning on $\{Z_1, Z_2\}$.
(b) Both $W_1$ and $W_2$ are instruments. The circles at the end of
the edges indicate that the direction between $W_1$ and $W_2$ is
irrelevant, as well as the possibility of unmeasured confounding
among $\{W_1, W_2, X\}$. (c) The typical covariance constraints
(``tetrads'') that are implied by instrumental variables also happen
in the case where no instruments exist, implying that rank constraints
in the covariance matrix are not enough information to discover IVs.
(d) A case that is difficult even when considering information
from non-Gaussian distributions.}
\label{fig:several_ivs}
\end{figure*}

It is not possible to test whether some observable variable is an IV
from its joint distribution with $X$ and $Y$, but IV assumptions can
be tested under a variety of assumptions by exploiting constraints in
the joint distribution of multiple observable variables
\citep{tianjiao:01,brito:02,kuroki:05}. However, existing
contributions on parameter identification do not immediately translate to
discovery algorithms. Our contribution are two IV discovery algorithms: a
theoretical one, which is {\it complete} (in a sense to be made
precise) with respect to a widely used graphical characterization of
IVs; and a practical one, which although might not be complete,
provides a practical alternative to the existing methods as the set of
assumptions required is fundamentally different.

The structure of the paper is as follows. In Section
\ref{sec:background}, we discuss basic concepts of IV modeling, and the
current state-of-the-art. We assume prior exposure to causal graphical
models and structural equation models \citep{sgs:00,pearl:00,bol:89},
including common concepts in causal graphical models such as
d-separation, back-door paths, colliders and active paths\footnote{A
partial summary for convenience: a vertex $V$ is {\it active on a path}
with respect to a conditioning set $\mathbf S$ if it is (i) a collider
in this path and itself or one of its descendants is in $\mathbf S$;
OR (ii) not a collider and not in $\mathbf S$. A path is {\it active}
if all of its vertices are active, {\it blocked} otherwise.  A path
between some $V_i$ and $V_j$ is {\it into} $V_i$ if the edge adjacent
to $V_i$ in this path points to $V_i$. A {\it back-door} (path)
between $V_i$ and $V_j$ is a path without colliders that is into $V_i$
and $V_j$.}. In Section \ref{sec:theory}, we discuss the theory behind
an IV discovery algorithm that is ``complete'' according to some
equivalence class of models.  The resulting algorithm has several
practical issues, and a more realistic alternative is provided in
Section
\ref{sec:algorithms}, which is then validated experimentally in
Section \ref{sec:experiments}.

\section{BACKGROUND}
\label{sec:background}

We assume a linear DAG causal model with observable
variables $\mathbf V \cup \{X, Y\}$. $X$ and $Y$ do not precede
any element of $\mathbf V$. $Y$ does not precede $X$. The goal is
to estimate the differential causal effect of $X$ on $Y$.

This task is common in applied sciences, as in many cases we have a
particular causal effect $X \rightarrow Y$ to be estimated, and a set
of covariates preceding $X$ and $Y$ is available. See
\cite{morgan:15} for several examples. This is in contrast to the more
familiar causal structure discovery tasks in the machine learning
literature, where an equivalence class of a whole causal system is
learned from data, and where some causal queries may or may not be
identifiable \citep{sgs:00}. The focus here in on quantifying the strength of a
particular causal query with background variables, as opposed to
unveiling the directionalities and connections of a causal graph. This 
allows more focused algorithms that bypass a full graph 
estimation. This philosophy has been exploited by \cite{entner:12} as a way of finding
possible sets of observable variables that can block the effect of any
hidden common cause of $X$ and $Y$. It does not, however, provide a causal
effect estimate if such a set does not exist.

When unmeasured confounding remains, the existence of a variable such
as $W$ in a system such as the one in Figure \ref{fig:iv} will provide
an alternative estimator. Using $\sigma_{ab.\mathbf s}$ to represent
the (conditional) covariance of two variables $A$ and $B$ (given set
$\mathbf S$), the parameterization in (\ref{eq:sem}) implies
$\sigma_{wx} = \lambda_{xw}\sigma_{ww}$, $\sigma_{wy} =
\lambda_{yx}\lambda_{xw}\sigma_{ww}$. It follows that
$\lambda_{yx} = \sigma_{wy} / \sigma_{wx}$. We can estimate
$\sigma_{wy}$ and $\sigma_{wx}$ from observations, allowing for a
consistent estimate of $\lambda_{yx}$. Notice that 
$\sigma_{wx} \neq 0$ is required.

A variable that is not an IV may be a {\it conditional} IV. This
means that if in the corresponding causal graph we find some set
$\mathbf Z$ that deactive some relevant paths, then we can identify
$\lambda_{yx}$ as $\sigma_{wy.\mathbf z} / \sigma_{wx.\mathbf
z}$. Figure \ref{fig:several_ivs}(a) illustrates a case. A graphical condition
for $W$ given $\mathbf Z$ is described by \cite{brito:02} as follows:
\begin{enumerate}[nolistsep]
\item $\mathbf Z$ does not d-separate $W$ from $X$ in $\mathcal G$;
\item $\mathbf Z$ d-separates $W$ from $Y$ in the graph obtained by
removing the edge $X \rightarrow Y$ from $\mathcal G$;
\item $\mathbf Z$ are non-descendants of $X$ and $Y$ in $\mathcal G$.
\end{enumerate}

For the rest of the paper, we will call the above condition the {\it graphical criteria
for instrumental variable validity}, or simply ``Graphical Criteria.''

The simple inference methods we just described required knowing the
causal graph. That being unavailable, the relevant structure needs to
be learned from the data. The lack of an edge in Figure \ref{fig:iv}
is not testable \citep{tianjiao:01}, but in a situation such as Figure
\ref{fig:several_ivs}(b), the {\it simultaneous} lack of edges $W_1
\rightarrow Y$ and $W_2 \rightarrow Y$ has a testable implication, as
in both cases we have $\lambda_{yx} = \sigma_{w_1y} / \sigma_{w_1x}$
and $\lambda_{yx} = \sigma_{w_2y} / \sigma_{w_2x}$. This leads to what
is known as a {\it tetrad constraint},
\begin{equation}
\sigma_{w_1y}\sigma_{w_2x} - \sigma_{w_1x}\sigma_{w_2y} = 0,
\end{equation}
\noindent which can be tested using observable data. Unfortunately, the tetrad constraint is
necessary, but not sufficient, to establish that both elements in this
pair of variables are instrumental.

Consider Figure \ref{fig:several_ivs}(c). It is not hard to show that
$\sigma_{w_1y.z}\sigma_{w_2x.z} - \sigma_{w_1x.z}\sigma_{w_2y.z} =
0$. However, the Graphical Criteria for IVs is not satisfied and,
indeed, $\lambda_{yx}$ can be vastly different from
$\sigma_{w_1y.z}/\sigma_{w_1x.z}$.  The core of our contribution is to
show how we can complement conditions such as tetrad constraints with
other conditions, tapping into the theory developed for linear
non-Gaussian causal discovery introduced in the LiNGAM framework of
\cite{shimizu:06}. 

\subsection{Previous Work}

\cite{hoyer:08b} describes a method for inferring linear causal effects among pairs
that are also confounded by hidden variables. The method, however,
requires large sample sizes and the knowledge of the number of hidden
common causes. Although finite, the number of possible differential
causal effects that are compatible with the data increases with the
number of assumed hidden variables. 

The use of tetrad constraints for testing the validity of particular
edge exclusions in linear causal models has a long history, dating
back at least to \cite{spearman:1904}.  More recently, it has been used in the
discovery of latent variable model structure
\citep{sil:06,spirtes:13}, where structures such as Figure \ref{fig:several_ivs}(c)
emerge but no direct relationships among observables (such as $X
\rightarrow Y$) are discoverable. The combination of tetrad constraints and
non-Gaussianity assumptions has been exploited by \cite{shimizu:09},
again with the target being relationships among latent variables. Tetrad
tests for the validity of postulated IVs were discussed by
\cite{kuroki:05}. The literature on learning algorithms allowing for
latent variables has been growing steadily for a long time,
including the Fast Causal Inference algorithm of \cite{sgs:00} and
more recent methods that exploit constraints other than independence
constraints \citep{tashiro:14,nowzohour:15}, but none of these methods
allow for the estimation of the causal effect of $X$ and $Y$ when
there is an unblocked unmeasured confounder between them. \cite{hoover:13}
introduced an algorithm for IV discovery, but it does not take into
account unidentifiability issues that can be solved by exploring
constraints other than covariance matrix constraints. Moreover, it attempts
to recover a much complex graph than that is necessary to solve this
particular question.

\begin{figure}[top]
\begin{center}
\includegraphics[height=3.1cm]{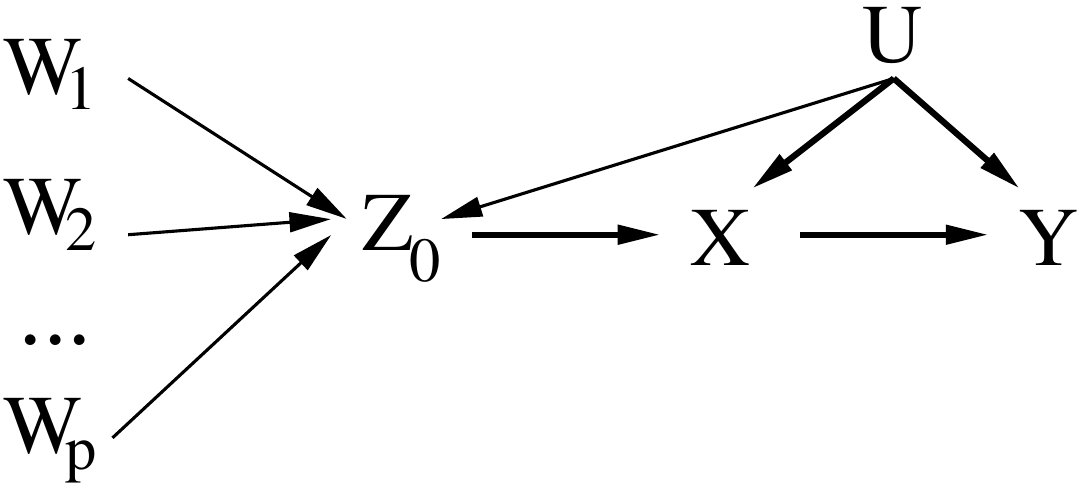}
\end{center}
\caption{In this model, variables $W_1, \dots W_p$ are instrumental variables conditioning on the empty set.
However, conditioning on $Z_0$ will introduce an active path from each
$W_i$ to $Y$ via $U$, destroying their validity.  This is particularly
an issue for algorithms such as {\tt sisVIVE} \citep{kang:15}, where a
variable is either deemed an IV or a conditioning variable.}
\label{fig:sisvive_collapses}
\end{figure}

\subsection{Directions}

A recent algorithm for the discovery of instrumental variables has
been introduced by \cite{kang:15}. It sidesteps the problems
introduced by models such as the one in Figure
\ref{fig:several_ivs}(c) by a clever choice of assumptions: it is
assumed that at least half of $\mathbf V$ are ``valid'' IVs, by which
this means that we can partition $\mathbf V$ into two sets, $\mathbf V
= \mathbf W \cup \mathbf Z$, such that each $W \in \mathbf W$ is a
conditional IV given $\mathbf Z \cup \mathbf W \backslash \{W\}$. This is done without
knowledge of which variables are valid and which are not. There are
situations where this assumption is plausible, and the resulting
algorithm ({\tt sisVIVE}, ``some invalid, some valid IV estimator'')
is very elegant and computationally efficient.

However, it does not take much to invalidate this assumption even when
nearly all of $\mathbf V$ can be used as instruments. Consider Figure
\ref{fig:sisvive_collapses} where we have an arbitrary number of IVs
$W_1, \dots, W_p$ that are valid by conditioning on the empty
set. {\it None} of them are valid by conditioning on $Z_0$, and in
this situation {\tt sisVIVE} may perform badly. In the following
Sections, we introduce an alternative approach that exchanges the ``at
least half valid, given everybody else'' condition with a less stringent
condition on validity, combined with assumptions of non-Gaussianity
and variations of the faithfulness assumption used in common causal
discovery algorithms \citep{sgs:00}. In practice, however, we do
exploit {\tt sisVIVE} as a useful building block in a practical
algorithm in Section \ref{sec:algorithms}. One difficulty is that for
models such as in Figure \ref{fig:several_ivs}(d), we will still not
be able to directly reject $\{W_1, W_2\}$ as invalid and further
assumptions would be needed.

\begin{algorithm}[t]
 \SetKw{Next}{} 
 \SetKwInOut{Input}{input}
 \SetKwInOut{Output}{output}
 \Input{Jointly distributed zero-mean random variables $\mathbf V \cup \{X, Y\}$;}
 \Output{The differential causal effect of $X$ on $Y$}
 \BlankLine
  Let $\Sigma$ be the covariance matrix of $\mathbf V \cup \{X, Y\}$\\
 \For{each pair $\{W_i, W_j\} \subseteq \mathbf V$}{
  \For{every set $\mathbf Z \subseteq \mathbf V \backslash \{W_i, W_j\}$ of no decreasing size}{
    \If{$\sigma_{w_ix.\mathbf z} = 0$ {\bf or} $\sigma_{w_jx.\mathbf z} = 0$}{     
      {\bf next}
    }
    \If{$\sigma_{w_ix.\mathbf z}\sigma_{w_jy.\mathbf z} \neq \sigma_{w_iy.\mathbf z}\sigma_{w_jx.\mathbf z}$}{     
      {\bf next}
    }
    $r_{W_i} \leftarrow $ \FuncSty{resproj}$(W_i, $  \FuncSty{lmb}$(W_i, \mathbf Z \cup \{W_j\}))$\\
    $r_{W_j} \leftarrow $ \FuncSty{resproj}$(W_j, $  \FuncSty{lmb}$(W_j, \mathbf Z \cup \{W_i\}))$\\
    $r_{Y_i} \leftarrow $ \FuncSty{resproj}$(Y, $  \FuncSty{lmb}$(W_i, \mathbf Z \cup \{W_j\}))$\\
    $r_{Y_j} \leftarrow $ \FuncSty{resproj}$(Y, $  \FuncSty{lmb}$(W_j, \mathbf Z \cup \{W_i\}))$\\
    \If{$r_{W_i} \cind r_{Y_i}$ {\bf and} $r_{W_j} \cind r_{Y_j} $}{
     \Return{$\sigma_{w_iy.\mathbf z} / \sigma_{w_ix.\mathbf z}$}
    }
  }
 }
 \Return{{\tt NA}}
 \BlankLine
\caption{An algorithm that learns the causal effect of $X$ on $Y$ knowing the joint distribution of
  all observable random variables. {\tt resproj}$(V, \mathbf S)$ is a function that returns
  the residual of the least-squares projection of $V$ into row vector $\mathbf S$, that is,
  $V - \mathbf S \times E[\mathbf S^T \mathbf S]^{-1}E[\mathbf S^T V]$. Function {\tt lmb}$(V, \mathbf S)$
  returns a {\it local Markov blanket}: all variables $S \in \mathbf S$ which are not independent of $V$
  given $\mathbf S \backslash S$, which is testable.}
\label{tab:algo1}
\end{algorithm}

\section{THEORY}
\label{sec:theory}

We assume our causal model is a {\it LiNGAM model}, a linear structural
equation model with independent, non-Gaussian error terms, which may
include latent variables \citep{shimizu:06}. Some of the intermediate
results in this Section will not require non-Gaussianity.

Algorithm \ref{tab:algo1} provides a method for inferring the
causal effect of some given $X$ on some given $Y$, getting as
input the joint distribution of $\mathbf V \cup \{X, Y\}$. This is
equivalent to having an oracle that replies yes or no to questions on
particular tetrad and independence constraints. The goal is to show
how we can provably find the correct causal effect in the limit of
infinite data, or to say we cannot identify it (the ``{\tt NA}''
return value). However, analogous to \citep{hoyer:08b}, there is some
subtle but important equivalence class of results we need to
consider. An algorithm for learning causal effects from empirical data
is discussed in Section \ref{sec:algorithms}.

\subsection{Preliminaries}

Following \cite{spirtes:13}, we call a {\it rank constraint} in a
matrix $M$ any constraint of the type $rank(M) \leq r$, where $r$ is
some constant. If $M$ is the cross-covariance submatrix given by
variables $\{V_i, V_j\}$ indexing the rows, and $\{V_k, V_l\}$ indexing the
columns, then the rank constraint $rank(M) \leq 1$ implies
$\sigma_{ik}\sigma_{jl} - \sigma_{il}\sigma_{jk} = 0$, as the latter
is the determinant of $M$.

We will use this notion in tandem with {\it t-separation}
\citep{sullivant:10}. First, let a {\it trek} $T$ from $V_i$ to $V_j$ in a graph
be an ordered pair of (possibly empty) directed paths $(P_1; P_2)$
where: $P_1$ has {\it sink} (vertex without children in $T$) $V_i$;
$P_2$ has sink $V_j$; and $P_1, P_2$ have the same {\it source}
(vertex in $T$ without parents in $T$). The ordered pair of vertex sets
$(\mathbf C_I; \mathbf C_J)$ {\it t-separates} vertex set $\mathbf{V_I}$ from 
vertex set $\mathbf{V_J}$ if, for every trek $(P_1; P_2)$ from a vertex in $\mathbf{V_I}$ to a
 vertex in $\mathbf{V_J}$, either $P_1$ contains a
vertex in $\mathbf C_I$ or $P_2$ contains a vertex in $\mathbf
C_J$. See \cite{spirtes:13} and \cite{sullivant:10} for a
generalization of this notion and further examples.

One relevant example can be obtained from Figure \ref{fig:several_ivs}(b).
Here, $\mathbf C_I = \emptyset$ and $\mathbf C_J = \{X\}$; $\mathbf{V_I} = 
\{W_1, W_2\}, \mathbf{V_J} = \{X, Y\}$. In Figure \ref{fig:several_ivs}(c),
$\mathbf C_I = \emptyset$ and $\mathbf C_J = \{U_1, Z\}$; $\mathbf{V_I} = 
\{Z, W_1, W_2\}, \mathbf{V_J} = \{Z, X, Y\}$.

Let $\Sigma_{\mathbf A \mathbf B}$ be the cross-covariance matrix of set
$\mathbf A$ (rows) and set $\mathbf B$ (columns). The DAG Trek Separation Theorem of
\cite{sullivant:10} says:

\textbf{Theorem 1 (Trek Separation for DAGs).} 
\textit{Let $\mathcal G$ be a DAG with vertex set $\mathbf V$. 
Let $\mathbf A$ and $\mathbf B$ be subsets of $\mathbf V$.
We have $rank(\Sigma_{\mathbf A \mathbf B}) \leq r$ in all linear
structural equation models with graph $\mathcal G$ if and only if
there exist subsets $\mathbf C_A$ and $\mathbf C_B$ of $\mathbf V$
with $|\mathbf{C_A}| + |\mathbf{C_B}| \leq r$ such that
$(\mathbf{C_A}; \mathbf{C_B})$ t-separates $\mathbf A$ from $\mathbf
B$.}

To jump from (testable) rank constraints to (unobservable) constraints in $\mathcal G$,
we assume our model distribution $P$ is {\it linearly rank-faithful}
to a DAG $\mathcal G$ \citep{spirtes:13}: that is, every
rank-constraint holding on a covariance (sub)matrix derived from $P$
is entailed by every linear structural model Markov with respect to
$\mathcal G$ \citep{spirtes:13}. 
{\it Linear faithfulness}, the assumption that vanishing partial
correlations hold in the distribution if and only if a corresponding
d-separation also holds in $\mathcal G$ \citep{sgs:00}, is a special
case of rank faithfulness, as t-separation implies d-separation
\citep{sullivant:10}.


\subsection{The Role of Non-Gaussianity}

In the Graphical Criteria introduced in Section \ref{sec:background},
the challenging condition is the second, as the first is easily
testable by faithfulness and the third is given by assumption. Another
way of phrasing condition 2 is: 2a, there is no active (with respect
to $\mathbf Z$) back-door path between $W$ and $Y$, nor any active
path that includes a collider, that does not include $X$; 2b, there is
no active directed path from $W$ to $Y$ that does not include $X$.  In
the next Section, we will partially address 2b. Here, we exploit
non-Gaussianity assumptions to partially tackle 2a. Our proof 
holds ``almost everywhere,'' in the sense it holds for all but a
(Lebesgue) measure zero subset of the set of possible structural
coefficients $\Lambda_{\mathcal G} = \{\lambda_{ij}\ |\ V_j \in
par_{\mathcal G}(i)\}$.  

The motivation for this concept is analogous to the different
variations of faithfulness, see the discussion on generic
identifiability by \cite{foygel:11} and \cite{sullivant:10} for more background on excluding
vanishing polynomials that are not a function of the graphical
structure. For instance, the completeness of the do-calculus
\citep{ilya:06,huang:06} would be of limited relevance if in many
models there were other adjustments by conditioning and
marginalization that did not follow from the graphical structure.
More specifically, linear faithfulness also holds almost everywhere in 
linear DAG models and it is assumed implicitly.

The main result of this section is the following:

\textbf{Theorem 2.} \textit{Let $\mathbf V \cup \{Y\}$ be a set of variables in a 
zero-mean LiNGAM model where $Y$ has no descendants. For some $V_i \in \mathbf V$, let
$\mathbf Z$ be its local Markov blanket (all $S \in \mathbf V\backslash\{V_i\}$ that
are d-connected to $V_i$ given $\mathbf V\backslash\{S, V_i\}$). Let $r_i \equiv V_i -
\mathbf a^T \mathbf Z$ be the residual of the  least-squares regression of $V_i$ on $\mathbf Z$,
with $\mathbf a$ being the corresponding least-squares
coefficients. Analogously, let $r_y \equiv Y - b_i V_i - \mathbf b^T
\mathbf Z$ be the residual of the corresponding least-squares
regression. Then, almost everywhere, $r_i \cind r_y$ if and
only if there are no active (with respect to $\mathbf Z$) back-door
paths between $V_i$ and $Y$, nor any active path that includes a
collider.}

The proofs of this and the next result are given in the Supplementary Material.

\subsection{Equivalence Class Characterization}

Even when using non-Gaussianity, there are still structures which are indistinguishable
by Algorithm \ref{tab:algo1}. They form an equivalence class characterized as follows:

\textbf{Theorem 3 (Downstream Conditional Choke Point Equivalence Class).} 
\textit{Suppose the outcome of Algorithm \ref{tab:algo1}, found with
respect to some $\{W_i, W_j, \mathbf Z\}$, is not correct under rank
faithfulness and the assumptions of Theorem 2. Then, for each $W \in
\{W_i, W_j\}$: (i) there is a directed path from $W$ to $Y$ that is
not blocked by $\{X\} \cup \mathbf Z$; (ii) the possible common
ancestors of $W$ and elements in this path are blocked by $\mathbf Z$;
(iii) this path includes some $Z_0 \notin
\mathbf Z \cup \{W_i, W_j, X\}$, where all
directed paths from $W$ to $Y$ in $\mathcal G$ are blocked by $\{Z_0,
X\} \cup \mathbf Z$; (iv) all directed paths from $W$ to $X$ are
blocked by $\mathbf Z \cup \{Z_0\}$}.

The result is that any tuple $(W_i, W_j, \mathbf Z)$ that satisfies
the conditions used by Algorithm \ref{tab:algo1} in effect belongs to
an equivalence class of possible tuples, some of which may provide an
incorrect causal effect. The common graphical feature in this
equivalence class is what we call a ``downstream conditional choke
point,'' illustrated by vertex $U$ in Figure
\ref{fig:several_ivs}(d). The name ``downstream'' denotes that this
point is a descendant of $\{W_i, W_j\}$, and has no active back-door paths
with them. The name ``choke point'' is due to the fact that other
common causes might exist between $X$ and $Y$, but no active paths,
other than the directed paths passing through this choke point, will
exist between $Y$ and $\{W_i, W_j\}$\footnote{The
literature has characterizations of {\it unconditional} choke
points \citep{shafer:93, sullivant:10}, relating them to unconditional
tetrad constraints. To the best of our knowledge, this is the first
time that conditional choke points are explicitly defined and
used. This is not a direct reuse of previous results, as the DAG class
is not closed under conditioning \citep{richardson:02}. Previous results were derived either for DAGs
or for special cases of conditioning \citep{sullivant:10}.}.

\begin{algorithm}[t]
 \SetKw{Next}{} 
 \SetKwInOut{Input}{input}
 \SetKwInOut{Output}{output}
 \Input{Data sample $\mathcal D$ from the joint distribution of zero-mean random variables $\mathbf V \cup \{X, Y\}$;
        A threshold $T$, defining the size of the background set}
 \Output{An estimate of the differential causal effect of $X$ on $Y$}
 \BlankLine
  \For{$V_i \in \mathbf V$}{
     $score_i \leftarrow$ \FuncSty{resDependenceScore}$(\mathcal D, V_i, \mathbf V, Y)$
  }
  Let $\mathcal B$ be the top $T\%$ of $\mathbf V$, as ranked by $score$\\
  $(\mathbf W, \mathbf Z) \leftarrow$ {\tt sisVIVE}$(\mathcal D, \mathbf V \backslash \mathcal B, \mathcal B, X, Y)$\\
  $\mathcal C \leftarrow \mathcal B$\\
  \While{TRUE}{
     $v^\star \leftarrow argmax_{v \in \mathcal C}$
         \FuncSty{BScore}$(\mathcal D, \mathbf W, \mathbf Z, \mathcal C \backslash \{v\}, X, Y)$\\
     \eIf{\FuncSty{BScore}$(\dots, \mathcal C \backslash \{v^\star\}) > $
         \FuncSty{BScore}$(\dots, \mathcal C)$}{
       $\mathcal C \leftarrow \mathcal C \backslash \{v\}$
     }{
       \bf break
     }
  }
  $(\mathbf W, \mathbf Z, dce) \leftarrow$ {\tt sisVIVE}$(\mathcal D, \mathbf V \backslash \mathcal B, \mathcal C, X, Y)$\\
 \Return{{\tt dce}}
 \BlankLine
\caption{The algorithm uses dependence measures of residuals under non-Gaussianity assumptions to 
score which variables are most likely not appropriate as instrumental variables. It then runs {\tt sisVIVE}
in a subset of remaining candidates, with some refinement.  Function
{\tt resDependenceScore} returns a quantification of how strongly
associated $r_i$ and $r_y$ are after (Markov blanket) least-squares regression of $V_i$
on $\mathbf V \backslash \{V_i\}$, and $Y$ on $\mathbf V$. See main text for further explanations.}
\label{tab:algo2}
\end{algorithm}

By isolating this feature, we know how to explain the possible
disparities obtained by letting a modified Algorithm \ref{tab:algo1}
return the causal effects implied by each acceptable tuple. By knowing
there is a {\it single} choke point per pair, which is {\it
unconfounded} with the ``candidate IVs,'' {\it and} which lies on all
unblocked directed paths from $W_i$ to $X$, more sophisticated
algorithms, combined with background knowledge, can be constructed
that exploit this piece of information. However, different pairs might
have different choke points, and we leave the description of a more
complex algorithm for future work. If all tuples agree and we assume
there is at least one valid tuple where $\{W_i, W_j\}$ are indeed IVs
conditional on $\mathbf Z$, then we are done, as guaranteed by this
simple result:

\textbf{Theorem 4 (Completeness).} 
\textit{If there is a pair of observable variables $\{W_i, W_j\}$ are IVs conditioned on 
some $\mathbf Z$ according to the Graphical Criteria, and each $W \in
\{W_i, W_j\}$ has no active back-door path with $X$ given $\mathbf Z$,
nor any active path that includes a collider, then Algorithm
\ref{tab:algo1} will find one.}

\noindent \textbf{\textit{Proof of Theorem 4.}} The test in Step 4 will not 
reject any such pair by linear faithfulness, as by the Graphical Criteria,
$\mathbf Z$ d-connects the pair to $X$. The test in Step 14 will not reject
any such pair, since by Theorem 2 the test will reject the pair only if there is
an active back-door path or a collider path between $W$ and $Y$. These situations
are excluded by the Graphical Criteria, except in the case where such paths exist between
$W$ and $X$, as the concatenation of those with the edge $X \rightarrow Y$ would exclude
$W$ from consideration. $\Box$

To summarize: Algorithm \ref{tab:algo1} is sound, in the limit of
infinite data, if we assume no downstream conditional choke point
exists in the graph. A necessary but not sufficient test to falsify
this assumption is by allowing a exhaustive check of all tuples $(W_i,
W_j, \mathbf Z)$ with a minimal $\mathbf Z$, and verifying whether
they imply the same causal effect. The algorithm is complete in the case where
for at least on pair $\{W_i, W_j\}$ the conditioning set $\mathbf Z$
also blocks active back-door/collider paths into $X$. This means, for
example, that the algorithm will not find answers in models where $W$
and $X$ have common causes that cannot be blocked, even if $W$ is a
valid IV by not having common causes with $Y$. For example, $W$ is a
valid IV in the model with paths $W \rightarrow X \rightarrow Y$, $W
\leftarrow U_1 \rightarrow X
\leftarrow U_2 \rightarrow Y$, but $W$ will be discarded due to the
back-door path between $W$ and $Y$ that is unblocked by not
conditioning on $X$.

\section{CHALLENGES AND A PARTIAL SOLUTION}
\label{sec:algorithms}

There are two major issues with Algorithm \ref{tab:algo1}. First,
testing (conditional) tetrad constraints often lead to many
statistical errors, which can be mitigated by some elaborated tricks
to take into account the redundancy of some constraints
\citep{sil:06,spirtes:13}. This however leads to a complicated and not necessarily
robust method. Second, an exhaustive search is in general not
computationally feasible.

Instead, we combine ideas inspired by the theoretical findings of the
previous Section with ideas underlying {\tt sisVIVE} \citep{kang:15}.
One practical issue properly addressed by {\tt sisVIVE} is that
we want to discover as many (conditional) IVs as possible, as typically
they individually will be weakly associated with the outcome $Y$.

Algorithm \ref{tab:algo2} modifies {\tt sisVIVE} in the following way.
In lines 1-3, we score each variable $V_i$ by estimating its
least-squares residual $r_i \equiv V_i - \mathbf a^T \mathbf V_{-i}$,
where $\mathbf V_{-i}$ is the vector formed by the local Markov
Blanket of $V_i$ within all remaining variables in $\mathbf V$ (see
definition in Algorithm \ref{tab:algo1}). Least-squares residuals $r_{y_i}
\equiv Y - b_i V_i - \mathbf b_{-i}^T \mathbf V_{-i}$ are also estimated. We use a measure
of dependence between the two residuals to define $score_i$. In our
implementation, tested in the next Section, we used the negative of
the p-value of Hoeffding's independence test between $r_i$ and $r_y$,
but other measures such as the HSIC \citep{gretton:07} could be used
instead. The idea is to flag variables which might be linked to $Y$ by
``strong'' active back-door paths, as motivated by Theorem 2, by
marking a proportion of them (as given by parameter $T$) as unsuitable
IV candidates. In our experiments, the proportion is set to $50\%$.

Line 5 executes {\tt sisVIVE} for a preliminary run, where we
indicate: $X$ and $Y$, the treatment and outcome variables; a set
$\mathcal B$, background variables to condition on {\it but not to
consider as possible instrumental variables}; and a set
$\mathbf V \backslash \mathcal B$ which will be split into a set of
(``valid'') IVs $\mathbf W$ and a set of (``invalid'') conditioning
variables $\mathbf Z$. As discussed in Section \ref{sec:background}, a
weak point of {\tt sisVIVE} is the impossibility of discarding ``bad''
conditioning variables. At this stage, however, we are still
conditioning on all variables, but aiming at avoiding some of the most
catastrophic mistakes of including a strongly confounded variable
(given everything else) into the pool of IVs.

A refinement takes place in lines 6-14. We shrink a conditional set
$\mathcal C$ initialized with $\mathcal B$. Function
{\tt BScore} (``back-door score'') aims at measuring how ``strong''
paths are between elements of $\mathbf W$ and $Y$ that might go
through back-doors or by conditioning on a sequence of colliders that
ends in a back-door with $Y$. The actual implementation of {\tt
BScore}, used in the next Section, is simple: for each $W_i$, estimate
residual $r_i$ conditional on $\mathbf W_{-i} \cup \mathbf Z \cup
\mathcal C \backslash \{c\}$ and the p-value of its dependence with $r_y$ (defined by
least-squares of $Y$ on $\mathbf W \cup \mathbf Z \cup \mathcal C \backslash
\{c\}$). The score is the product of all p-values over the elements of
$\mathbf W$. We then shrink set $\mathcal C$ to a local optimal.
A last run of {\tt sisVIVE} is then performed, returning the estimated
differential causal effect {\it dce}.

We are aware of several shortcomings of this algorithm\footnote{It is a greedy
method, and bad local optimal might happen. For instance, once a
variable is excluded from the pool of possible IVs, it never goes
back. Parameter $T$ should be chosen in a way that we believe a
reasonable number of conditional IVs will remain.  There is no formal
guarantee that in an example such as Figure
\ref{fig:sisvive_collapses}, variable $Z_0$ will be ranked higher than
$W_i$ by {\tt resDependenceScore}, although heuristically this is
justified by dependences typically decaying as longer paths are
traversed in a graph.}. The algorithm still relies on the underlying assumptions of
{\tt sisVIVE}, but relaxing it to require $50\%$ or more of valid IVs
only on the subset $\mathbf V \backslash
\mathcal B$. All these concerns are valid, but our point is to provide
a reasonably simple algorithm that is justified by (i) how Theorem 2
provides a recipe to remove bad conditioning variables for
{\tt sisVive}; (ii) how Theorem 3 justifies relying on further
assumptions, adapted from {\tt sisVIVE}, as non-Gaussianity by itself
is shown not to rule out invalid structures. Complex, more refined,
algorithms will be object of future work.  For now, we will show
empirically how even a partial solution such as Algorithm
\ref{tab:algo2} can provide improvements over the state-of-the-art
method.

\begin{figure}[top]
\begin{center}
\includegraphics[height=3.0cm]{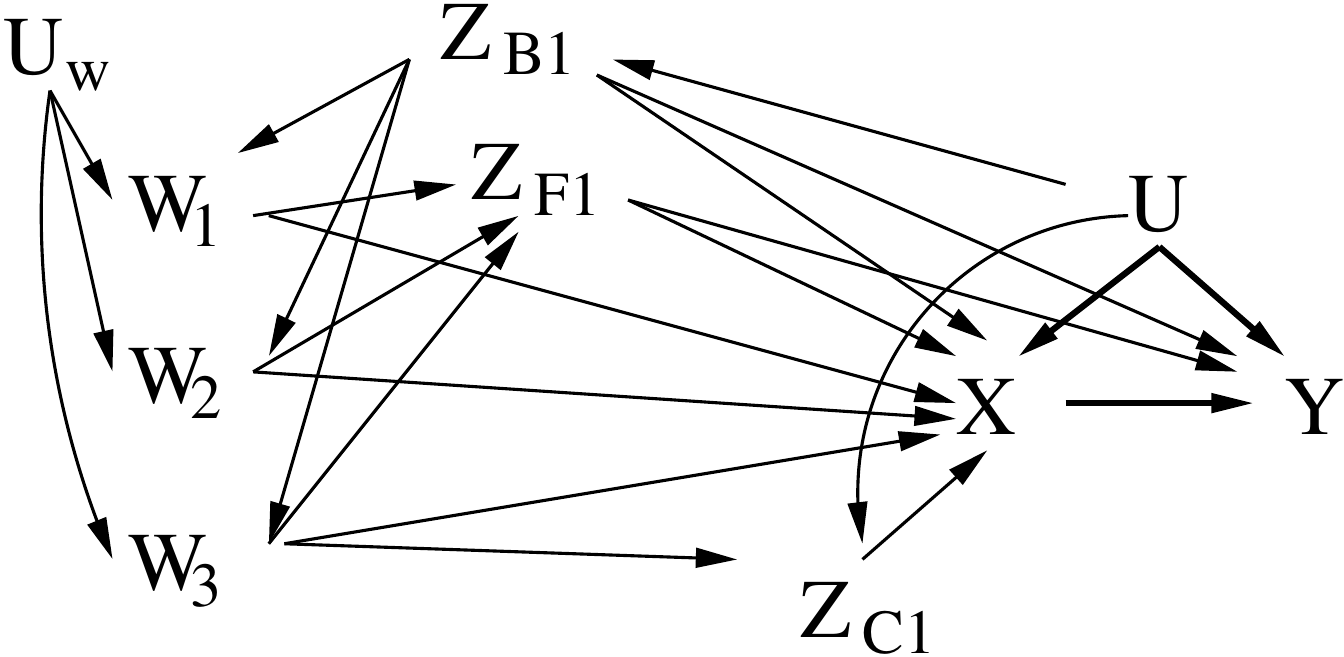}
\end{center}
\caption{Example of a synthetic graph generated by the template used in Section \ref{sec:experiments}.
$W_1, W_2$ and $W_3$ are valid IVs conditioned on $Z_{B1}$ and $Z_{F1}$ only, as
conditioning on $Z_{C1}$ activates the path $W_i \rightarrow Z_{C1} \leftarrow U \rightarrow Y$, which
invalidates the instruments.}
\label{fig:synth}
\end{figure}

\section{EXPERIMENTS}
\label{sec:experiments}

We assess how Algorithm \ref{tab:algo2} (which we will call {\tt B-sisVIVE}, 
as in ``back-door protected {\tt sisVIVE}'') compares to other methods
in a series of simulations.

\begin{table*}
\begin{center}
\begin{tabular}{r|ccc|ccc}
              & 100/0.25 & 1000/0.25 & 5000/0.25  & 100/0.50 & 1000/0.50 & 5000/0.50\\
\hline
NAIVE1        &  0.25    &   0.24    &   0.24     &  0.51    &   0.50    &  0.51     \\
NAIVE2        &  0.28    &   0.27    &   0.26     &  0.53    &   0.54    &  0.55     \\
NAIVE3        &  0.22    &   0.19    &   0.19     &  0.40    &   0.41    &  0.41     \\
\hline
ORACLE        &  0.12    &   0.03    &   0.01     &  0.22    &   0.05    &  0.02     \\
W-ORACLE      &  0.17    &   0.09    &   0.07     &  0.30    &   0.20    &  0.19     \\
S-ORACLE      &  0.18    &   0.07    &   0.03     &  0.36    &   0.14    &  0.05     \\
\hline
SISVIVE       &  0.20    &   0.13    &   0.15     &  0.41    &   0.38    &  0.42     \\
B-SISVIVE     &  0.16    &   0.16    &   0.12     &  0.38    &   0.33    &  0.28     \\
B-SISNAIVE    &  0.18    &   0.15    &   0.14     &  0.38    &   0.36    &  0.32     \\
\end{tabular}
\end{center}
\caption{Experimental results. Errors are measured by the median absolute difference
between the estimated causal effect and true $\lambda_{yx}$, over 200
trials in each experimental condition.}
\label{tab:experiments}
\end{table*}

The simulations are performed as follows: we generate synthetic graphs
by four groups of variables. Group $\mathbf W$ are variables which can
be used as conditional IVs. Group $\mathbf Z_D$ are variables which
lie on directed paths from $\mathbf W$ to $X$ and $Y$. Group $\mathbf
Z_C$ are variables which are children of $\mathbf W$ and $U$, the
unmeasured confounder between $X$ and $Y$.  Finally, group $\mathbf
Z_B$ are children of $U$ and parents of $\mathbf W$\footnote{More
precisely, the children of $\mathbf W$ are in $\{X\} \cup \mathbf Z_D
\cup \mathbf Z_C$ and its parents are $\mathbf Z_B$ and a second latent variable $U_w$. The children of
$\mathbf Z_F$ are $\{X, Y\}$, its parents are $\mathbf W$. The
children of $\mathbf Z_C$ is only $X$, its parents are $\{U\} \cup
\mathbf W$.  The children of $\mathbf Z_B$ are $\{X, Y\} \cup \mathbf
W$ and its parent is only $U$.}. Figure \ref{fig:synth} shows an
example where $|\mathbf W| = 3, |\mathbf Z_F| = 1, |\mathbf Z_C| = 1,
|\mathbf Z_B| = 1$.

The methods we compare against are: {\sc NAIVE1}, obtained by
least-squares of $Y$ on $X$, assuming no confounding; {\sc NAIVE2},
two-stage least squares (TSLS) using all variables as instruments; {\sc NAIVE3}, regression on
$X$ and all other variables, assuming no confounding; {\sc ORACLE},
using TSLS on the right set of IVs and adjustment set; {\sc
W-ORACLE}, uses $\mathbf W$ as IVs, but conditions on all of the other
variables; {\sc S-ORACLE}, {\tt sisVIVE} performed by first correctly
removing the set $\mathbf Z_C$; {\sc SISVIVE}, the \cite{kang:15} algorithm
taking all variables as input; {\sc B-SISVIVE}, our method, with the same input;
{\sc B-SISNAIVE}, a variation of Algorithm \ref{tab:algo2}, by skipping
steps 7-14.

All error variables and latent variables are zero-mean Laplacian
distributed, and coefficients are sampled from Gaussians, such that the
observed variables have a variance of 1. Models are rejected until the
causal effect $\lambda_{yx}$ has an absolute value of 0.05 or
more. Coefficients $\lambda_{xu} = \lambda_{yu}$ are fixed at two
levels, $(0.25, 0.50)$, the higher the harder, as this makes
unmeasured confounding stronger. Sample sizes are set at 100, 1000,
5000. Comparisons are shown in Table \ref{tab:experiments}, with the
setup $|\mathbf W| = 25$, $|\mathbf Z_F| = 10$, $|\mathbf Z_B| = 1$,
$|\mathbf Z_C| = 10$. This satisfies the criterion of $|\mathbf W|$ being
more than the number of remaining variables, although only the 11 variables
$\mathbf Z_F \cup \mathbf Z_B$ should be used.

The message in Table \ref{tab:experiments} seems clear. In particular,
increasing the amount of confounding can make the problem considerably
harder; {\tt sisVIVE} works very well under the correct assumptions
(as seen by the performance of {\sc S-ORACLE}, which is just {\tt
sisVIVE} given the -- usually unknown -- information about which
variables one should not condition on for validating the possible
instrumental variables); otherwise, it can perform poorly ({\sc
SISVIVE}, which does hardly better than some na\"{i}ve approaches);
our method ({\sc B-SISVIVE}) can provide some sizeable improvements
over this state-of-the-art method. This is true even in its more
straightforward variation, which does not refine its choice of
conditioning set but only forbids some variables to be selected as
instruments ({\sc B-SISNAIVE}). We conclude these are important lessons
in the estimation of  causal effects with observational
data.

\section{CONCLUSION}
\label{sec:conclusion}

Finding instrumental variables is one of the most fundamental problems
in causal inference. To the best of our knowledge, this paper provides
the first treatment on how this can be systematically achieved by
exploiting non-Gaussianity and clarifying to which extent an equivalence class of
solutions remains. We then proceeded to show how non-Gaussianity can be
exploited in a pragmatic way, by adapting a state-of-the-art
algorithm. Finally, we illustrated how improvement can be considerable under
some conditions.

We expect that theoretical challenges in instrumental variable
discovery can be further tackled by building on the findings shown
here. In particular, as also hinted by \cite{kang:15}, some of the
ideas here raised extend to non-linear (additive) and binary
models. Methods developed in \cite{peters:14} can potentially
provide a starting point on how to allow for non-linearities in the
context of instrumental variables.

More sophisticated graphical criteria for the identification of causal
effects in linear systems were introduced by \cite{brito:02}. Further
work has led to rich graphical criteria to identify causal effects in
confounded pairs \citep{foygel:11}. This goes far beyond the standard
IV criteria discussed in Section \ref{sec:background}. It also opens
up the possibility of more elaborated discovery algorithms where
back-door blocking \citep{entner:12} and the methods in this paper
cannot provide a solution, but how to perform this task in a
computationally and statistically tractable way remains an open
question.

Code for the procedure and to generate synthetic studies is 
available at \url{http://www.homepages.ucl.ac.uk/~ucgtrbd/code/iv_discovery}.


\begin{thebibliography}{28}
\providecommand{\natexlab}[1]{#1}
\providecommand{\url}[1]{\texttt{#1}}
\expandafter\ifx\csname urlstyle\endcsname\relax
  \providecommand{\doi}[1]{doi: #1}\else
  \providecommand{\doi}{doi: \begingroup \urlstyle{rm}\Url}\fi

\bibitem[Bollen(1989)]{bol:89}
K.~Bollen.
\newblock \emph{Structural {E}quations with {L}atent {V}ariables}.
\newblock John Wiley \& Sons, 1989.

\bibitem[Brito and Pearl(2002)]{brito:02}
C.~Brito and J.~Pearl.
\newblock Generalized instrumental variables.
\newblock \emph{Proceedings of 18th Conference on Uncertainty in Artificial
  Intelligence}, 2002.

\bibitem[Chu et~al.(2001)Chu, Scheines, and Spirtes]{tianjiao:01}
T.~Chu, R.~Scheines, and P.~Spirtes.
\newblock Semi-instrumental variables: a test for instrument admissibility.
\newblock \emph{Proceedings of the Seventeenth Conference on Uncertainty in
  Artificial Intelligence (UAI 2001)}, pages 83--90, 2001.

\bibitem[Darmois(1953)]{darmois:53}
G.~Darmois.
\newblock Analyse g\'{e}n\'{e}rale des liaisons stochastiques.
\newblock \emph{Review of the International Statistical Institute},
  21:\penalty0 2--8, 1953.

\bibitem[Entner et~al.(2012)Entner, Hoyer, and Spirtes]{entner:12}
D.~Entner, P.O. Hoyer, and P.~Spirtes.
\newblock Statistical test for consistent estimation of causal effects in
  linear non-{G}aussian models.
\newblock \emph{Proceedings of the 15th International Conference on Artificial
  Intelligence and Statistics (AISTATS 2012)}, pages 364--372, 2012.

\bibitem[Foygel et~al.(2011)Foygel, Draisma, and Drton]{foygel:11}
R.~Foygel, J.~Draisma, and M.~Drton.
\newblock Half-trek criterion for generic identifiability of linear structural
  equation models.
\newblock \emph{Annals of Statistics}, 40:\penalty0 1682--1713, 2011.

\bibitem[Gretton et~al.(2007)Gretton, Fukumizu, Teo, Song, Sch\"{o}lkopf, and
  Smola]{gretton:07}
A.~Gretton, K.~Fukumizu, C.~Teo, L.~Song, B.~Sch\"{o}lkopf, and A.~Smola.
\newblock A kernel statistical test of independence.
\newblock \emph{Advances in Neural Information Processing Systems},
  20:\penalty0 585--592, 2007.

\bibitem[Hoyer et~al.(2008)Hoyer, Shimizu, Kerminen, and Palviainen]{hoyer:08b}
P.~Hoyer, S.~Shimizu, A.~Kerminen, and M.~Palviainen.
\newblock Estimation of causal effects using linear non-{G}aussian causal
  models with hidden variables.
\newblock \emph{International Journal of Approximate Reasoning}, 49:\penalty0
  362--378, 2008.

\bibitem[Huang and Valtorta(2006)]{huang:06}
Y.~Huang and M.~Valtorta.
\newblock Pearl's calculus of intervention is complete.
\newblock \emph{Proceedings of the Twenty-Second Conference on Uncertainty in
  Artificial Intelligence (UAI 2006)}, pages 217--224, 2006.

\bibitem[Kang et~al.(2015)Kang, Zhang, Cai, and Small]{kang:15}
H.~Kang, A.~Zhang, T.~Cai, and D.~Small.
\newblock Instrumental variables estimation with some invalid instruments and
  its application to mendelian randomization.
\newblock \emph{Journal of the American Statistical Association}, page To
  appear, 2015.

\bibitem[Kuroki and Cai(2005)]{kuroki:05}
M.~Kuroki and Z.~Cai.
\newblock Instrumental variable tests for directed acyclic graph models.
\newblock \emph{Tenth workshop on Artificial Intelligence and Statistics
  (AISTATS 2005)}, 2005.

\bibitem[Morgan and Winship(2015)]{morgan:15}
S.~Morgan and C.~Winship.
\newblock \emph{Counterfactuals and Causal Inference: Methods and Principles
  for Social Research}.
\newblock Cambridge University Press, 2015.

\bibitem[Nowzohour et~al.(2015)Nowzohour, Maathuis, and
  B\"{u}hlmann]{nowzohour:15}
C.~Nowzohour, M.~Maathuis, and P.~B\"{u}hlmann.
\newblock Structure learning with bow-free acyclic path diagrams.
\newblock \emph{arXiv:1508.01717}, 2015.

\bibitem[Pearl(2000)]{pearl:00}
J.~Pearl.
\newblock \emph{Causality: {M}odels, {R}easoning and {I}nference}.
\newblock Cambridge University Press, 2000.

\bibitem[Peters et~al.(2014)Peters, Mooij, Janzing, and
  Sch{\"o}lkopf]{peters:14}
J.~Peters, J.~M. Mooij, D.~Janzing, and B.~Sch{\"o}lkopf.
\newblock Causal discovery with continuous additive noise models.
\newblock \emph{Journal of Machine Learning Research}, 15:\penalty0 2009--2053,
  2014.

\bibitem[Phiromswad and Hoover(2013)]{hoover:13}
P.~Phiromswad and K.~Hoover.
\newblock Selecting instrumental variables: A graph-theoretic approach.
\newblock \emph{Working paper. Available at SSRN:
  http://ssrn.com/abstract=2318552 or http://dx.doi.org/10.2139/ssrn.2318552},
  2013.

\bibitem[Richardson and Spirtes(2002)]{richardson:02}
T.~Richardson and P.~Spirtes.
\newblock Ancestral graph {M}arkov models.
\newblock \emph{Annals of Statistics}, 30:\penalty0 962--1030, 2002.

\bibitem[Shafer et~al.(1993)Shafer, Kogan, and P.Spirtes]{shafer:93}
G.~Shafer, A.~Kogan, and P.Spirtes.
\newblock Generalization of the tetrad representation theorem.
\newblock \emph{DIMACS Technical Report}, 1993.

\bibitem[Shimizu et~al.(2006)Shimizu, Hoyer, Hyv\"{a}rinen, and
  Kerminen]{shimizu:06}
S.~Shimizu, P.~Hoyer, A.~Hyv\"{a}rinen, and Antti Kerminen.
\newblock A linear non-gaussian acyclic model for causal discovery.
\newblock \emph{Journal of Machine Learning Research}, 7:\penalty0 2003--2030,
  2006.

\bibitem[Shimizu et~al.(2009)Shimizu, Hoyer, and Hyv\"{a}rinen]{shimizu:09}
S.~Shimizu, P.~Hoyer, and A.~Hyv\"{a}rinen.
\newblock Estimation of linear non-gaussian acyclic models for latent factors.
\newblock \emph{Neurocomputing}, 72:\penalty0 2024--2027, 2009.

\bibitem[Shpitser and Pearl(2006)]{ilya:06}
I.~Shpitser and J.~Pearl.
\newblock Identification of conditional interventional distribution.
\newblock \emph{Proceedings of the Twenty-Second Conference on Uncertainty in
  Artificial Intelligence (UAI 2006)}, pages 437--444, 2006.

\bibitem[Silva et~al.(2006)Silva, Scheines, Glymour, and Spirtes]{sil:06}
R.~Silva, R.~Scheines, C.~Glymour, and P.~Spirtes.
\newblock Learning the structure of linear latent variable models.
\newblock \emph{Journal of Machine Learning Research}, 7:\penalty0 191--246,
  2006.

\bibitem[Skitovitch(1953)]{skitovitch:53}
W.~Skitovitch.
\newblock On a property of the normal distribution.
\newblock \emph{Doklady Akademii Nauk SSSR}, 89:\penalty0 217â??--219, 1953.

\bibitem[Spearman(1904)]{spearman:1904}
C.~Spearman.
\newblock ``{G}eneral intelligence,'' objectively determined and measured.
\newblock \emph{American Journal of Psychology}, 15:\penalty0 210--293, 1904.

\bibitem[Spirtes(2013)]{spirtes:13}
P.~Spirtes.
\newblock Calculation of entailed rank constraints in partially non-linear and
  cyclic models.
\newblock \emph{Proceedings of the Twenty-Ninth Conference on Uncertainty in
  Artificial Intelligence (UAI 2013)}, pages 606--615, 2013.

\bibitem[Spirtes et~al.(2000)Spirtes, Glymour, and Scheines]{sgs:00}
P.~Spirtes, C.~Glymour, and R.~Scheines.
\newblock \emph{Causation, {P}rediction and {S}earch}.
\newblock Cambridge University Press, 2000.

\bibitem[Sullivant et~al.(2010)Sullivant, Talaska, and Draisma]{sullivant:10}
S.~Sullivant, K.~Talaska, and J.~Draisma.
\newblock Trek separation for {G}aussian graphical models.
\newblock \emph{Annals of Statistics}, 38:\penalty0 1665--1685, 2010.

\bibitem[Tashiro et~al.(2014)Tashiro, Shimizu, Hyv\"{a}rinen, and
  Washio]{tashiro:14}
T.~Tashiro, S.~Shimizu, A.~Hyv\"{a}rinen, and T.~Washio.
\newblock {ParceLiNGAM}: A causal ordering method robust against latent
  confounders.
\newblock \emph{Neural Computation}, 26:\penalty0 57--83, 2014.

\end{thebibliography}

\section*{APPENDIX: Supplementary Material}

We present here proofs of Theorems 2 and 3. The result for Theorem 2
depends on this standard theorem \citep{darmois:53,skitovitch:53}:

\textbf{Theorem 5. (Darmois-Skitovitch Theorem)} \textit{Let $e_1, \dots, e_n$ be independent
random variables, $n \geq 2$. Let $v_1 = \sum_i \alpha_i e_i$, $v_2 =
\sum_i \beta_i e_i$ for some coefficients $\{\alpha_i\}$,
$\{\beta_i\}$. If $v_1$ and $v_2$ are independent, then those $e_j$ for
which $\alpha_j \neq 0$, $\beta_j \neq 0$ are Gaussian}.

The idea is that if we assume $\{e_i\}$ are not Gaussian, $\{V_1, V_2\}$
share a common source if and only if they are dependent. See
\citep{shimizu:06,entner:12} for a deeper discussion on how this
theorem is used in causal discovery.

For the main results, we will assume particular algebraic (polynomial) identities
implied by the model graph do not vanish at the particular parameter
values of the given model (which we called ``almost everywhere'' results
in the theorem). We will in particular consider ways of
``expanding'' the structural equations of each vertex according to
{\it exogenous} variables, that is, any variable which is either an
error term or latent variable (assuming without loss of generality
that latent variables have no parents).

For each vertex $V_k$ in the model, and each exogenous ancestor $E_m$
of $V_k$, let $\mathcal P_{km}$ be the set of all directed paths from
$E_m$ to $V_k$. For each path $p \in \mathcal P_{km}$, define
$\phi_{kmp} =
\prod_j \lambda_{jj'}$, the product of all coefficients along this
path for $V_j \in \mathbf V \cap p$ where $V_{j'} = p \cap
par_{\mathcal G}(j)$ (that is, $V_{j'}$ is the parent of $V_j$ in this path. We multiply coefficients following a sequence
$E_m \rightarrow \dots \rightarrow V_{j'} \rightarrow V_j \rightarrow \dots \rightarrow V_k$). From this,
\begin{equation}
  V_k = \sum_{E_m \in \mathcal A_{\mathcal G}(k)} \sum_{p \in \mathcal P_{km}} \phi_{kmp}E_m,
 \label{eq:xpand_exo}
\end{equation}
\noindent where $A_{\mathcal G}(k)$ is the set of exogenous ancestors of $V_k$,
where for $E_m = e_k$ we have $\phi_{kmp} \equiv 1$ and path $p$ is
given by the single edge $e_k \rightarrow V_k$. We refer to the idea of
expansion a few times in the proofs as a way of describing how the
models can be written as polynomial functions of the coefficients
$\Lambda_{\mathcal G} = \{\lambda_{ij}\ |\ V_j \in par_{\mathcal
G}(i)\}$.

Overall, for a LiNGAM model $\mathcal M$ with DAG $\mathcal G$, we
denote by $\mathcal X_{\mathcal G}$ the set of exogenous variables of
$\mathcal M$, and by the {\it expanded graph} of $\mathcal M$ the
graph $\mathcal G$ augmented with the error terms and the corresponding
edges $e_i \rightarrow V_i$ for all observable vertices $V_i$ in $\mathcal G$.

The main result used in the proof of Theorem 2 comes from the
following Lemma. Notice that the non-Gaussianity assumption and the
Darmois-Skitovitch Theorem are not necessary for its proof.

\textbf{Lemma 6.} 
\textit{Let $\mathbf V \cup \mathbf U$ be the set of variables in a 
zero-mean LiNGAM model $\mathcal M$, where $\mathbf U$ are the
latent variables of the model. For some $V_i \in \mathbf V$, let
$\mathbf V_{\backslash i}$ be $\mathbf V \backslash \{V_i\}$.  Let
$r_i \equiv V_i - \mathbf a^T \mathbf V_{\backslash i}$ be the
residual of the least-squares regression of $V_i$ on $\mathbf
V_{\backslash i}$, with $\mathbf a$ being the corresponding
least-squares coefficients. Then, almost everywhere, $r_i$
can be written as a linear function of the exogeneous variables of
$\mathcal M$, $r_i = \sum_{E_m \in \mathcal X_{\mathcal G}} c_m E_m$,
where $c_m \neq 0$ if and only if $V_i$ is d-connected to $E_m$ given
$\mathbf V \backslash V_i$ in the expanded graph of $\mathcal M$.}

\noindent \textbf{\textit{Proof of Lemma 6.}} Without loss of generality, assume
that each latent variable in $\mathbf U$ has no parents. We will sometimes use
$X_k$ as another representation of any particular model variable
(observable, latent or error term), with the index $k$ indicating
particular variables in $\mathbf V \cup \mathbf U$ and error terms,
depending on the context.

One way of obtaining $r_i$ is by first performing least-squares
regression of each model variable $X_k$ on $V_j$, for some $V_j \neq
V_i$ in $\mathbf V$, and calculating residuals $X_k^{(1)}$.  Define
$\mathbf V^{(1)}$ as the set of all residuals $\{V_k^{(1)}\}$, $k \neq
j$. We then repeat the process by regressing on some element of
$\mathbf V^{(1)}\backslash \{V_i^{(1)}\}$, iterating until we are left
with $\mathbf V^{(n - 1)}$ containing the single element $V_i^{(n - 1)}$, where $n$ is the size of
$\mathbf V$ and $V_i^{(n - 1)} = r_i$. The elimination sequence can be
arbitrary.

Let $V_j$ be a vertex in $\mathbf V_{\backslash i}$.  Let
$\lambda_{km}$ be the structural coefficient between $V_k$ and any
$X_m \in \mathbf{V} \cup \mathbf{U}$. We define $\lambda_{kj} \equiv 0$ if $X_j$ is not a parent of
$V_k$. Since
\[
V_k = \lambda_{kj}V_j + \sum_{X_m \in par_{\mathcal G}(k)\backslash V_j} \lambda_{km}X_m + e_k,
\]
\noindent we have
\[
\sigma_{kj} = \lambda_{kj}\sigma_{jj} + \sum_{X_m \in par_{\mathcal G}(k) \backslash V_j} \lambda_{km}\sigma_{mj} +
 \sigma_{e_kj},
\]
\noindent where $\sigma_{e_kj}$ is the covariance of $e_k$ and $V_j$ and $\sigma_{mj}$ here
represents the covariance of $X_m$ and $V_j$. This implies,
\begin{equation}
a_{kj}^{(1)} = \lambda_{kj} + \sum_{X_m \in par_{\mathcal G}(k) \backslash V_j}\lambda_{km}a_{j}^{(1)} + a_{e_kj}^{(1)}.
\label{eq:a_relation}
\end{equation}
\noindent where $a_{e_kj}^{(1)}$ is the least-squares regression coefficient of $e_k$ on $V_j$.
This means $V_k^{(1)} = V_k - a_{kj}^{(1)}V_j$ can be written as
\begin{equation}
V_k^{(1)} = \sum_{X_m \in par_{\mathcal G}(k) \backslash V_j}\lambda_{km}X_{m}^{(1)} + e_k^{(1)}
\label{eq:a_relation}
\end{equation}
\noindent with $X_m^{(1)}$ and $e_k^{(1)}$ defined analogously.

We can iterate this process until we are left with $r_i$:
\begin{equation}
r_i = \sum_{U_k \in par_{\mathcal G}(i) \cap \mathbf  U}\lambda_{ik}U_k^{(n - 1)} + e_i^{(n - 1)},
\label{eq:a_relation2}
\end{equation}
\noindent where $|\mathbf V| = n$. Variable $U_k^{(n - 1)}$ is the residual of the regression of $U_k$ on
$\mathbf V_{\backslash i}$, similarly for $e_i^{(n - 1)}$.

What we will show next is that within (\ref{eq:a_relation2}) each
$U_k^{(n - 1)}$ and $e_i^{(n - 1)}$ can be expanded as polynomial
functions of $\Lambda_{\mathcal G}$ and $\mathcal X_\mathcal G$, and
the end result will contain non-vanishing monomials that are a (linear)
function of only the exogenous variables $E_m$ which are d-connected
to $V_i$ given $\mathbf V\backslash V_i$ in the expanded graph of
$\mathcal M$. Since the monomials cannot vanish except for a strict
subset of lower dimensionality than that of the set of possible
$\Lambda_\mathcal G$, the result will hold almost everywhere.

Since we are free to choose the elimination ordering leading to $r_i$,
as they all lead to the same equivalent relation
(\ref{eq:a_relation2}), let us define it in a way that a vertex can be
eliminated at stage $t$ only when its has no ancestors in $\mathbf
V^{(t - 1)}$ (where $\mathbf V^{(0)} \equiv \mathbf V_{\backslash
i}$).

For $t = 1$, the only exogenous variables which will have a non-zero
coefficient multiplying $V_j$ in the least-squares regression are the
parents of $V_j$ in the expanded graph, since $V_j$ has no other
ancestors\footnote{Assuming $V_j$ is not a child of $V_i$. In this
case, without loss of generality we assume that the parents of $V_i$ are added
to the parents of $V_j$, and remove $V_i$ from the model at any
iteration $t$.}.  Let $U_k^{(1)}$ be the residual of some latent
parent of $V_j$,
\begin{equation}
\displaystyle
  U_k^{(1)} = U_k - a_{kj}\left(\sum_{E_m \in \mathcal A_{\mathcal G}(j)} 
                           \sum_{p \in \mathcal P_{jm}} \phi_{jmp}E_m\right),
\label{eq:xpand_U}
\end{equation}
\noindent where $\phi_{jmp} \equiv \lambda_{jm}$ if $E_m$ is a latent variable,
or 1 if $E_m = e_j$. Moreover, $a_{kj} = \lambda_{jk}v_{kk} / \sigma_{jj}$, where
$v_{kk}$ is the variance parameter of $U$ and $\sigma_{jj}$ is a polynomial function
of $\Lambda_\mathcal G$. We can multiply both sides of the equation above by 
$\sigma_{jj}$ (as well all equations referring to any $V_k^{(1)}$ or $X_m^{(1)}$ such as
(\ref{eq:a_relation})) to get a new system of variables that is polynomial in
$\Lambda_\mathcal G$. We will adopt this step implicitly and claim that from
(\ref{eq:xpand_U}) we have that $U_k^{(1)}$ can be expanded as parameters that
are polynomial functions of $\Lambda_\mathcal G$. Moreover, it is clear from
(\ref{eq:xpand_U}) that there will be at least one non-vanishing
monomial containing each $E_m$. In what follows, we refer to any expression analogous to
(\ref{eq:xpand_U}) as the {\it expansion} of $U_k^{(t)}$ for $t = 1, 2, \dots, n - 1$.

We define a DAG $\mathcal G^{(1)}$ with vertices $X_m^{(1)}$, where
$U_k^{(1)}$ will assume as children all and only the $V_k^{(1)}$ such that $U_k$ is
a parent of $V_k$ in the original extended graph of the model. That is,
$\mathcal G^{(1)}$ is the extended graph over residuals after the first
regression. The respective model $\mathcal M^{(1)}$ is given by equations
of type (\ref{eq:a_relation}) with parameters coming from 
$\Lambda_\mathcal G$\footnote{To be more precise, polynomial functions of such parameters,
as we are implicitly multiplying each equation by $\sigma_{jj}$.}.

For any $t > 1$, let $V_j^{(t)}$ be the vertex being eliminated.  Each
$U_k^{(t)}$ in which $U_k^{(t - 1)}$ is a parent of $V_j^{(t - 1)}$ in
$\mathcal G^{(t - 1)}$ will be a polynomial function of
$\Lambda_\mathcal G$ and a linear function the union of the exogenous
variables present in the expansion of each parent of $V_j^{(t - 1)}$:
the expansion analogous to (\ref{eq:xpand_U}) in the new model will
always introduce new symbols $\lambda_{j\star}$ into existing
monomials, or create new monomials with $e_j$, as vertex $V_j^{(t -
1)}$ had no eliminated descendants up to iteration $t$. As such, no
exogenous variable will be eliminated from the algebraic expansion of
the respective $U_k^{(t)}$.

Finally, the expansion of $\lambda_{ik}U_k^{(n - 1)}$ in
(\ref{eq:a_relation2}) will not cancel any monomial in the expansion
of some other $\lambda_{ik'}U_{k'}^{(n - 1)}$: since $U_k$ and
$U_{k'}$ are both parents of $V_i$, no monomial in the expansion of $U_k$
can differ from a monomial in the expansion of $U_{k'}$ by a factor
of $\lambda_{ik}\lambda_{ik'}$. So (\ref{eq:a_relation2}) will depend
algebraically on the union of the exogenous terms leading to each
$U_k^{(n - 1)}$.

To prove the Lemma, we start by pointing out that $U_k^{(n - 1)}$ will
have a latent/error parent of some $V_j$ in its expansion if and only
if there is at least one sequence of vertices $(V_c, \dots, V_j)$
where $V_c$ is an observable child of $U_k$ and any two consecutive
elements in this sequence have at least one common latent parent in
$\mathcal G$ (the sequence can be a singleton, $V_c = V_j$). To see
this, notice that the different $U_k^{(t)}$ form an equivalence
relation: each $U_k^{(t)}$ with a $V_j^{(t - 1)}$ child which is being eliminated at iteration $t$ will
include into its expansion the exogenous variables found in the
expansion of the other parents of $V_j^{(t - 1)}$. This partitions $\mathbf
V_{\backslash i}$ into sets in which each vertex $V_j$ can ``reach''
some other vertex $V_k$ by first moving to some $V_{j'}$ which shares
a latent parent with $V_j$ and which can ``reach'' $V_k$. The latent parents
of $\mathbf V$ are then partitioned according to their observed children.

To finalize the proof, suppose $V_i$ is d-separated from a
latent/error parent $E_m$ of $V_j$ given $\mathbf V \backslash
V_i$. This happens if and only if all latent parents of $V_i$ (and
$e_i$) are d-separated from $E_m$ given $\mathbf V
\backslash V_i$. Let $U_k$ be a latent parent of $V_i$ (or its error term). 
Then $U_k^{(n - 1)}$ cannot have $E_m$ in its expansion. If this was the
case, by the previous paragraph $U_k$ would be d-connected to all
latent parents of $V_j$, meaning $V_i$ would be d-connected to
them. This implies $c_m = 0$. Conversely, suppose $V_i$ is d-connected
to the error term or a latent parents of $V_j$ given $\mathbf V
\backslash V_i$. Then again by the previous paragraph, for any latent
parent $U_k$ of $V_i$, $U_k^{(n - 1)}$ will have the latent parents of
$V_j$ as terms in its expansion, implying $c_m \neq
0$ almost everywhere. $\Box$

We can now prove Theorem 2.

\noindent \textbf{\textit{Proof of Theorem 2.}} Considering the system for
$\{V_i, Y\} \cup \mathbf Z$, we can represent the model in an
equivalent way where all latent variables are exogenous. Applying
Lemma 6 to both $r_i$ and $r_y$, and by Theorem 5, these variables
will be dependent if and only if they are a non-trivial linear
function of at least one common exogeneous variable $E_m$ in the
model. By Lemma 6, this happens if and only if $V_i$ is d-connected to
$E_m$ given $\mathbf Z$ and $Y$ is d-connected to $E_m$ given $V_i$
and $\mathbf Z$. If $Y$ is d-connected to $E_m$ given $\mathbf Z$
only, and since the concatenation of the $(V_i, E_m)$ path with $(Y,
E_m)$ path must be by either colliding at the same child of $E_m$, or
connected through some $V_x \leftarrow E_m \rightarrow V_y$, where
$V_x$ is in the path connected to $V_i$ (which needs to be into $V_x$)
and $V_y$ is in the path connect to $Y$ (which is into $V_y$), the
theorem holds.  If $Y$ is not d-connected to $E_m$ given $\mathbf Z$
only, then $Y$ must be d-connected to $V_i$ given $\mathbf Z$ by a
path that is into $V_i$, and the claim again follows. $\Box$

The proof of the final result is as follows:

\noindent \textbf{\textit{Proof of Theorem 3.}} Point (i): according to the
Graphical Criteria for IVs, there should be an active path from $W$
to $Y$ that does not include $X$, or otherwise the algorithm would
return the right answer. This path has to be directed, as any other
possible active path based on back-doors or conditioning on colliders
has been ruled out by the non-Gaussian residual test (Theorem 2).

Point (ii) is related: if there was an active back-door path
between $W$ and some $Z_0$ connected to $Y$ by an active directed path, then
the concatenation of the two paths would lead to an active back-door path
between $W$ and $Y$, contrary to the result of the residual test.

Now we show point (iii). Since $\sigma_{w_ix.\mathbf z}\sigma_{w_jy.\mathbf z} -
\sigma_{w_iy.\mathbf z}\sigma_{w_jx.\mathbf z} = 0$, the covariance
submatrix formed by using $(\mathbf Z, W_i, W_j)$ as rows and
$(\mathbf Z, X, Y)$ as columns has determinant $|\Sigma_{\mathbf
Z\mathbf Z}||\sigma_{w_ix.\mathbf z}\sigma_{w_jy.\mathbf z} -
\sigma_{w_iy.\mathbf z}\sigma_{w_jx.\mathbf z}| = 0$, which follows from
standard block matrix decompositions. Rank faithfulness, combined with
the Trek Separation Theorem, and the fact that $\mathbf Z$ is of
minimal size (i.e., there is no proper subset of it satisfying the
tests in Algorithm \ref{tab:algo1}), implies there is a pair of sets
$(\mathbf C_{\mathbf Z \cup \{W_i, W_j\}}, \mathbf C_{\mathbf Z \cup
\{X, Y\}})$ such that the rank of the covariance submatrix $(\mathbf
Z, W_i, W_j) \times (\mathbf Z, X, Y)$ is $|\mathbf Z| + 1$.  There
are (trivial) treks of zero edges between elements of $\mathbf Z$,
implying all of $\mathbf Z$ is necessary for the
t-separation to hold. Moreover, as $\mathbf Z$ is minimal, it is not possible
for a vertex $Z$ in $\mathbf Z$ to be both in $\mathbf C_{\mathbf Z
\cup \{W_i, W_j\}}$ and $\mathbf C_{\mathbf Z \cup \{X, Y\}}$, as this
would create an active path from $W$ to $Y$ colliding at $Z$, contrary
to the tests in the algorithm.  Therefore, there is exactly one other
vertex not in $\mathbf Z$ needed for the t-separation to hold. Then any treks
from $W$ to $Y$ will have to go through this vertex, which by point (ii)
have to be directed paths. 

For point (iv): if there is an unblocked path from $W$ to $X$ not going
through $\mathbf Z \cup \{Z_0\}$, this would contradict that 
$|(\mathbf C_{\mathbf Z \cup \{W_i, W_j\}}, \mathbf C_{\mathbf Z \cup
\{X, Y\}})| = |\mathbf Z| + 1$, as there would be no elements left to
cover this extra trek. $\Box$

{\bf Remarks:} The assumptions are stronger than, for instance, the
ones used in the proofs of \cite{tashiro:14}. A closely related result
in that paper is its Lemma 2, a result identifying the dependence
between the residual of the regression of a variable on its
children. It does not use any variation of the faithfulness
assumption. This is because, in their context, it is enough to detect
the dependence between the residual and {\it some} children. So if
some path cancellations take place, some other path cancellations
cannot occur. But we need the dependence of our $r_i$ and every
relevant error term, because we cannot claim that $r_i$ depends on
{\it some} error terms or latent variables, while $r_y$ depends on
{\it some} error terms or latent variables, if these two sets do not
overlap. Although some of the ideas by \cite{tashiro:14} could be used
in our context to build partial models and from the deduce
instrumental variables, it goes against our framework of solving a
particular prediction problem (causal effect of a target
treatment-outcome) directly, instead of doing it by recovering parts
of a broader causal graph.

Finally, we have not provided an explicit discussion on how to
validate the non-Gaussianity assumption by testing the non-Gaussianity
of the residuals, as done by \cite{entner:12}. Or, more precisely, showing
which assumptions are necessary so that testing non-Gaussianity of the
residuals is equivalent to testing non-Gaussianity of the error
terms. This is left as future work.

\end{document}